# Enhancing Environmental Monitoring through Multispectral Imaging: The WasteMS Dataset for Semantic Segmentation of Lakeside Waste

Qinfeng Zhu[1,2], Ningxin Weng[1], Lei Fan[1](✉), and Yuanzhi Cai[3]

[1] Department of Civil Engineering, Xi'an Jiaotong-Liverpool University, Suzhou, 215123, China
Lei.Fan@xjtlu.edu.cn
[2] Department of Computer Science, University of Liverpool, Liverpool, L69 3BX, UK
[3] CSIRO Mineral Resources, Kensington, WA 6151, Australia

**Abstract.** Environmental monitoring of lakeside green areas is crucial for environmental protection. Compared to manual inspections, computer vision technologies offer a more efficient solution when deployed on-site. Multispectral imaging provides diverse information about objects under different spectrums, aiding in the differentiation between waste and lakeside lawn environments. This study introduces WasteMS, the first multispectral dataset established for the semantic segmentation of lakeside waste. WasteMS includes a diverse range of waste types in lawn environments, captured under various lighting conditions. We implemented a rigorous annotation process to label waste in images. Representative semantic segmentation frameworks were used to evaluate segmentation accuracy using WasteMS. Challenges encountered when using WasteMS for segmenting waste on lakeside lawns were discussed. The WasteMS dataset is available at https://github.com/zhuqinfeng1999/WasteMS.

## 1 Introduction

Compared to Red-Green-Blue (RGB) imaging methods, multispectral imaging captures reflections from the real world in multiple specific bands, providing richer information than visible light alone [1]. Multispectral imaging is vital for remote sensing, agriculture, and environmental monitoring, as the infrared band offers feature perception capabilities that visible light cannot [2]. By analyzing specific infrared bands, multispectral imaging can reveal material composition, vegetation health, and other environmental parameters. Thus, in complex outdoor environments with varying lighting conditions, multispectral information may enhance the perception capabilities of a model in remote sensing tasks. As such, this approach has been widely adopted in remote sensing. Currently, most remote sensing satellites are equipped with both RGB and multispectral or hyperspectral imaging cameras, and numerous studies have demonstrated the importance of multispectral imaging analysis [3].

Despite the proven potential of multispectral imaging in various vision tasks, it has not been applied to the environmental monitoring of lakesides [4], specifically in perceiving waste. As transitional zones between land and water, lakesides present unique challenges due to varying lighting conditions, diverse vegetation debris, and complex backgrounds. Relying solely on RGB information for semantic segmentation tasks can be challenging [5]. Furthermore, because lakesides are popular for activities like camping, barbecuing, and gatherings, large amounts of various types of waste are often left behind, posing additional challenges for environmental monitoring [6].

Identifying lakeside waste using multispectral images can readily be achieved through semantic segmentation. Semantic segmentation is a core task in computer vision, aiming at assigning a semantic category label to each pixel in an image [7]. This task is widely applied in various fields, including autonomous driving, medical imaging [8], remote sensing [9, 10], and environmental monitoring [11], where precise pixel-level interpretation is crucial.

For enhanced segmentation accuracy, semantic segmentation is often conducted using a deep learning approach, which often relies on neural networks and dedicated training datasets. On the network side, commonly used neural networks, such as Convolutional Neural Networks (CNNs) and Vision Transformers (ViTs) [12], have demonstrated their effectiveness in semantic segmentation [13]. Recently, autoregressive networks such as Mamba [14-16], RWKV [17, 18], and xLSTM [19-21] have achieved even higher efficiency and accuracy in segmenting high-resolution images [22]. On the training-dataset side, existing multispectral datasets are primarily focused on controlled environments such as agriculture, while datasets targeting lakeside waste are scarce. This scarcity hinders the development of accurate waste perception and intelligent monitoring methods for lakeside environments.

Recognizing this gap, we introduce WasteMS, a multispectral dataset specifically designed for the semantic segmentation of lakeside waste. We used a multispectral camera to collect data on various types of lakeside waste under different weather conditions and at different times. The annotations were conducted through a rigorous process to ensure accuracy. Extensive testing was performed using representative semantic segmentation frameworks, and the benchmarking results of WasteMS were presented.

The contributions of this paper can be summarized as follows:
1. We present WasteMS, the first multispectral dataset for the semantic segmentation of lakeside waste, marking the initial effort to use multispectral information for lakeside waste monitoring.
2. We provide benchmark performance using representative semantic segmentation methods and discuss various encountered challenges.

## 2    The WasteMS Dataset

The WasteMS dataset consists of a total of 117 nine-channel multispectral images with a resolution of 682×682 pixels, along with their semantic segmentation annotations. We divided the dataset into training, validation, and test sets in an approximate ratio of 7:1:2 for standard deep learning training and evaluation of semantic segmentation. The training set contains 81 images, the validation set contains 11 images, and the test set contains 25 images. Examples of WasteMS and its annotations are shown in Fig. 1.

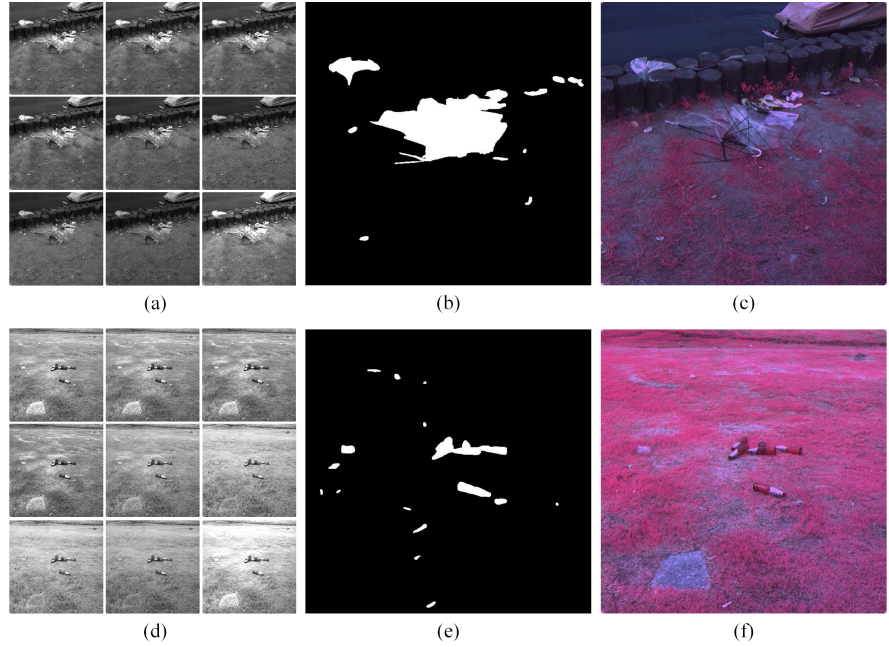

**Fig. 1.** Example images from the WasteMS dataset. (a) and (d) show the 9-channel data of two scenes; (b) and (e) are the ground truth mask; (c) and (f) are pseudo-color images composed using 9 channels.

### 2.1 Data Collection

We used the CMS4 multispectral camera for data collection, which covers a wavelength range that includes the visible red-light region as well as a rich near-infrared region. This wavelength range is well-suited for environmental monitoring and vegetation analysis [23]. Specifically, the CMS4 multispectral camera can capture information across 9 bands, including 8 narrowband color filters and one black-and-white filter. The central wavelengths ($\lambda$) of the 8 bands are 653nm, 695nm, 731nm, 772nm, 809nm, 851nm, 886nm, and 929nm, with maximum transmittance ($T_{max}$) ranging from 50% to 60%. Additionally, the camera includes a neutral density filter (Band 9), which operates over the spectral range of 500 to 1000nm, with an average transmittance ($T_{mean}$) of 12%.

Transmittance is an important factor in multispectral cameras, as it determines the amount of light at specific wavelengths. Higher transmittance means capturing more light. To clearly demonstrate the characteristics of the multispectral data we collected, the relationship between transmittance and the nine bands is illustrated in the Fig. 2. Each macro-pixel of the camera consists of a 3×3 pixel matrix, representing the 9 bands, resulting in an original data resolution of 2048×2048 pixels. After preprocessing, the 9 channels data can be acquired with a resolution of 682×682 pixels.

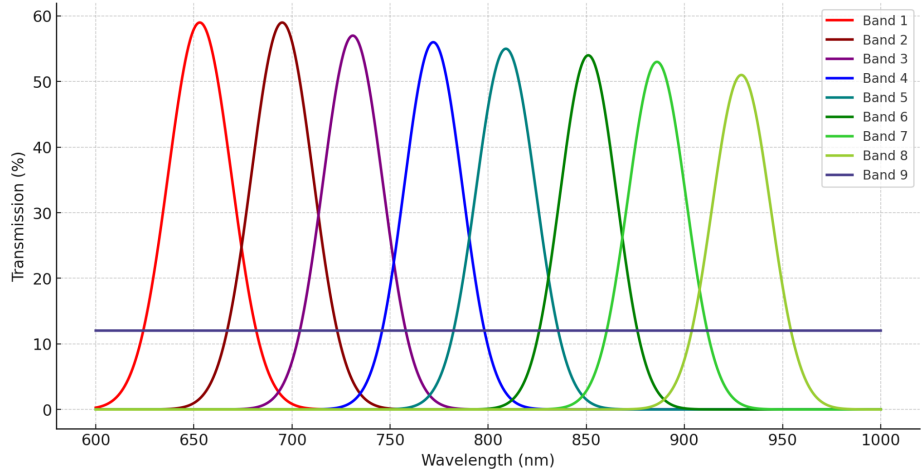

**Fig. 2.** The relationship between Wavelength and Transmission of the 9 channels obtained by the multispectral camera.

We conducted data collection on the lakeside lawns in Suzhou, China. These areas are frequently visited by tourists for activities such as camping, barbecuing, and gatherings, leading to the presence of significant amounts of waste. To ensure a diverse dataset, we collected data under various lighting conditions, including sunny, cloudy, and overcast weather. The data collection times ranged from noon to sunset, ensuring a variety of lighting angles. Furthermore, we collected various types of waste, including plastic bags, cans, cigarette butts, tissues, plastic bottles, plastic boxes, and cardboard. To ensure scene diversity, we also varied the locations, minimizing the number of repeated areas, and included a few complex scenes to enhance the dataset's robustness.

To further ensure the accuracy of the annotations, we not only collected multispectral images of the waste but also captured high-resolution RGB images using a camera. The purpose of this was to use high-resolution camera images for reference and verification when encountering difficult-to-distinguish targets, such as differentiating between leaves and paper. This approach also ensured that small targets, such as cigarette butts or small pieces of paper, were not missed.

### 2.2   Annotation

The annotation was carried out by multiple trained persons. During annotation, we fused the information from the 9 channels to create pseudo-color images and used annotation software Label Studio to annotate these pseudo-color images, as shown in Figure 3(a). To ensure the accuracy of the annotation outlines, we first used the Segment Anything Model (SAM) [24] to assist with the annotation. By inputting rectangular prompts, SAM's ViT pretrained model would infer and generate masks. In the second step, we manually adjusted these masks at the pixel level to ensure the precision of the mask edges.

In the process of annotation, high-resolution RGB images captured by the camera, as shown in Fig. 3(c), are used for reference to ensure the accuracy of annotation. After we completed the annotation, we implemented a review process to conduct secondary checks on the annotations to ensure their accuracy. The final masks are shown in Figure 3(b).

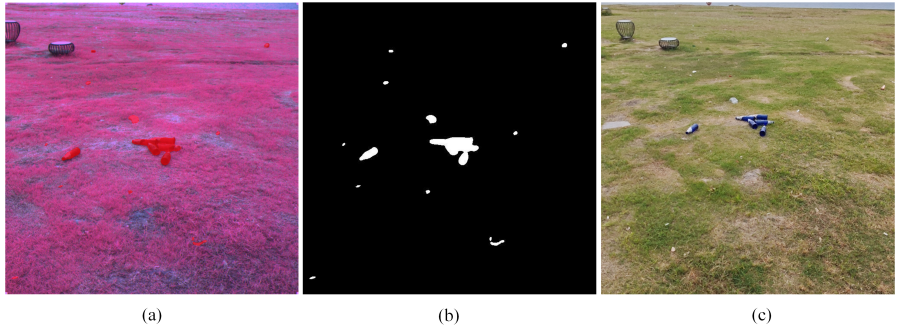

(a)                                   (b)                                   (c)

**Fig. 3.** Annotation diagram of WasteMS. (a) the labeling process is carried out using pseudo-color graphs, (b) the ground truth mask obtained after the labeling is completed, and (c) the high-resolution images for reference in the labeling process.

## 3        Benchmarks

### 3.1        Representative Baselines

We used several representative semantic segmentation methods to benchmark the WasteMS dataset. These methods adopt an encoder-decoder architecture [25], where the encoder benefits from pre-training on large image classification datasets like ImageNet [26, 27], enhancing its feature extraction capabilities.

For the encoder, we employed ResNet [28], ConvNeXt [29], and Swin Transformer[30]. ResNet introduces residual learning into the network to avoid vanishing and exploding gradient problems, allowing deeper networks to be effectively trained. ConvNeXt also utilizes the concept of residual connections but makes more significant internal structural improvements, such as using larger convolutional kernels, simplified normalization, and activation processes. This gives it a more powerful ability to handle image details, making it particularly effective in fine-grained semantic segmentation tasks. Swin Transformer, built on the hierarchical Vision Transformer, introduces shifted windows, which divides the image into multiple small patches and applies self-attention within these patches individually. This approach improves processing efficiency while capturing global dependencies across windows, making it better suited for handling large variations in waste scales at lakesides.

For the decoder, we employed Fully Convolutional Networks (FCN) [31], Pyramid Scene Parsing Network (PSPNet) [32], DeepLabV3+ [33, 34], and UperNet [35]. FCN is a pioneer in semantic segmentation tasks, replacing traditional fully connected layers with convolutional layers, allowing the network to accept input images of any size and output segmentation results of the same size. FCN uses transposed convolutions for upsampling, gradually restoring the image size, and is highly efficient in simple scenes. PSPNet introduces a pyramid pooling module to capture contextual information at multiple scales, integrating global scene understanding without losing details, making it better suited for complex scenes. DeepLabV3+ uses dilated convolutions to expand the receptive field of feature maps, capturing broader contextual semantic information. UperNet combines multiple structural features, not only employing multi-scale fusion techniques but also integrating pyramid pooling and full convolutional characteristics. This enables the network to effectively handle multi-scale information and optimize both global and local information.

### 3.2   Experimental Settings

When combining encoders and decoders, we used the combinations of Swin Transformer with UperNet, ConvNeXt with UperNet, ResNet with DeepLabV3+, ResNet with PSPNet, ResNet with UperNet, and ResNet with FCN as baselines to benchmark WasteMS. The experimental settings for these combinations are shown in Table 1. Among these, the combinations of ConvNeXt with UperNet and Swin Transformer with UperNet used the AdamW optimizer with a 1500-iteration warm-up strategy, while the other methods used the SGD optimizer without warm-up. The learning rates were chosen based on extensive experimentation to find the optimal values. Given the limited size of the WasteMS dataset, data augmentation was employed to effectively enhance generalization ability [36]. The augmentation strategies included random resize, random crop, random flip, and random rotate. All experiments were conducted using two RTX 4090D GPUs.

**Table 1.** Training settings for semantic segmentation networks on the WasteMS Dataset.

| Dataset | WasteMS |
|---|---|
| Resize | 512×512 |
| Patch size | 512×512 |
| Total training | 15000 iterations |
| Batch size | 16 |
| Optimizer | SGD/AdamW |
| Schedule | PolyLR |
| Loss function | Cross entropy loss |

### 3.3   Benchmarking Results

Table 2 presents the performance of various representative semantic segmentation frameworks on the WasteMS test set, including fine-tuning results after pre-training on ImageNet1k and fully supervised training results. We used the Intersection over Union (IoU) metric to evaluate the segmentation results. Additionally, we calculated the number of parameters (Parms) and the computational load (FLOPs) for each semantic segmentation network using a single 512×512 resolution, nine-channel multispectral image.

Among these, the combination of RseNet50 and UperNet with ImageNet pre-training achieved the highest IoU score. By comparing the results with and without pre-trained parameters, we observed that although methods showed better performance on the IoU metric after loading pre-trained parameters, the improvement was limited. This is due to the domain gap between the ImageNet dataset and WasteMS, primarily in terms of image data channels. Specifically, ImageNet images are RGB three-channel images of common objects, while WasteMS consists of nine-channel multispectral image data. This indicates that pre-training on RGB images may have a limited impact on downstream tasks involving multispectral data due to domain differences in the number of channels.

**Table 2.** The benchmark results of the representative segmentation methods. "w/o" represents "without pretraining". "w" represents "with pretraining".

| Encoder | Decoder | Pretrain | IoU(%) | #Parms(M) | #Flops(G) |
|---|---|---|---|---|---|
| ResNet50 | FCN | w/o | 40.30 | 47.12 | 198 |
|  |  | w | 43.62 |  |  |
| ResNet50 | PSPNet | w/o | 44.59 | 46.60 | 179 |
|  |  | w | 45.51 |  |  |
| ResNet50 | DeepLabV3+ | w/o | 53.22 | 41.22 | 177 |
|  |  | w | 55.03 |  |  |
| ResNet50 | UperNet | w/o | 54.88 | 64.04 | 237 |
|  |  | w | 61.77 |  |  |
| ConvNeXt-T | UperNet | w/o | 57.51 | 59.25 | 234 |
|  |  | w | 59.63 |  |  |
| Swin-T | UperNet | w/o | 57.63 | 58.95 | 236 |
|  |  | w | 57.93 |  |  |

To visualize the results, we show the segmentation results of each representative method after pre-training fine-tuning in Fig. 4. It can be observed that most methods are able to successfully segment larger waste, but they still encounter false negative errors when dealing with small, elongated objects, such as wooden sticks. This may be because the spectral characteristics of wooden sticks are similar to those of thin branches. Most methods are able to handle challenging scenarios with many small pieces of paper effectively, with the combination of ResNet and UperNet performing better and exhibiting fewer false positive errors.

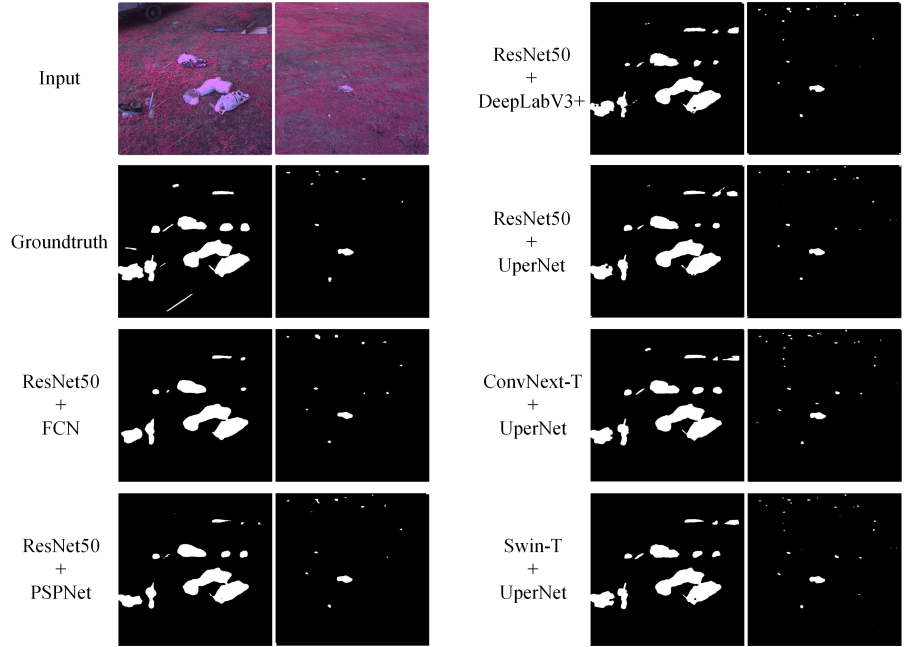

**Fig. 4.** Visual comparisons of segmentation outcomes of representative baselines on WasteMS.

## 4     Challenges and Outlook

### 4.1    Small Targets

In WasteMS, there is a significant variation in the scale of waste, with many small targets [37] present. During annotation, we referenced high-resolution camera images to ensure that very small objects, such as cigarette butts and small pieces of paper, were annotated, even though they constitute a very small proportion in the multispectral images. Although existing semantic segmentation networks can effectively segment larger waste items, such as plastic bottles and cans, they still face challenges with small objects like cigarette butts and small pieces of paper, often resulting in false negatives. In the future, it is important to develop networks that can effectively perceive these small objects, with a focus on more efficient fusion and feature extraction of multispectral information.

### 4.2    Pre-training Generalization

Pre-training and fine-tuning is an effective paradigm in semantic segmentation tasks [38], widely applied in remote sensing, medical imaging, and autonomous driving. However, as shown in Table 2, the improvement in performance on WasteMS from pre-training on ImageNet is limited because of the domain gap between WasteMS and ImageNet. Therefore, it is worth exploring more suitable pre-training strategies.

### 4.3   Channel Selection

WasteMS has 9 data channels, and an excessive number of channels may lead to data redundancy, increasing computational load and affecting segmentation accuracy. Thus, in future multispectral semantic segmentation networks, it is meaningful to explore channel selection strategies [39]. This can also improve the interpretability of multispectral data understanding, clearly quantifying the importance of each channel.

### 4.4   Limited Samples

Due to the limited scenarios of lakeside waste and the high cost of data collection and annotation, the number of WasteMS data samples is limited. Therefore, exploring data augmentation strategies for multispectral images in the future is worthwhile. Image generation technologies based on Generative Adversarial Networks (GANs) [40] and diffusion models [41] have been widely applied in recent years and have shown good results in augmenting image [42] and point cloud data [36]. However, research on data augmentation for multispectral images based on generative models is still limited and deserves further exploration. Additionally, developing more effective small-sample learning networks for multispectral data is a future research direction.

## 5   Conclusion

In this study, we introduced WasteMS, the first multispectral dataset tailored for semantic segmentation of lakeside waste, aiming at enhancing environmental monitoring through the utilization of multispectral imaging. The dataset encompasses a wide range of waste types under various lighting conditions, and has been carefully annotated to ensure high annotation accuracy. We conducted extensive testing using representative semantic segmentation networks, providing benchmark performance. We discussed in detail the challenges presented by the WasteMS dataset, including small objects, pre-training generalization, channel selection, and small sample size, and proposed future work to address these challenges.

The WasteMS Dataset for Semantic Segmentation of Lakeside Waste        11